\documentclass{article}
\usepackage[colorlinks=true,linkcolor=magenta,citecolor=green]{hyperref}
\usepackage[nonatbib, preprint]{neurips_2021}
\usepackage[utf8]{inputenc} 
\usepackage[T1]{fontenc}    
\usepackage{url}            
\usepackage{array,multirow,tabularx,ragged2e,booktabs}       
\usepackage{amsfonts}       
\usepackage{nicefrac}       
\usepackage{microtype}      
\usepackage{multicol}
\usepackage{colortbl}
\usepackage{pifont}
\usepackage{wrapfig}
\usepackage{subfig}
\newcommand{\xmark}{\ding{55}}%
\usepackage{enumitem}
\usepackage{physics}
\usepackage{wrapfig}
\usepackage{nicematrix}

\usepackage[toc,page,header]{appendix}
\usepackage{minitoc}
\numberwithin{equation}{section}
\usepackage[font=small,labelfont=bf]{caption}

\usepackage[backend=bibtex,style=ieee]{biblatex}
\usepackage{amsmath}
\usepackage{amssymb}
\usepackage{amsthm}
\usepackage{amsfonts}
\usepackage{mathtools}
\usepackage{bm}
\usepackage{cancel}
\usepackage{thmtools, thm-restate}
\usepackage{empheq}
\usepackage{pgfplots}
\usepackage{tikz}

\usepackage[frozencache, cachedir=./minted]{minted}
\usepackage[most]{tcolorbox}
\tcbuselibrary{minted,breakable,xparse,skins}

\definecolor{bg}{gray}{0.95}
\DeclareTCBListing{mintedbox}{O{}m!O{}}{%
  breakable=true,
  listing engine=minted,
  listing only,
  minted language=#2,
  minted style=default,
  minted options={%
    linenos,
    gobble=0,
    breaklines=true,
    breakafter=,,
    fontsize=\small,
    numbersep=8pt,
    #1},
  boxsep=0pt,
  left skip=0pt,
  right skip=0pt,
  left=25pt,  right=0pt,
  top=3pt,
  bottom=3pt,
  arc=5pt,
  leftrule=0pt,
  rightrule=0pt,
  bottomrule=2pt,
  toprule=2pt,
  colback=bg,
  colframe=blue!70!white,
  enhanced,
  overlay={%
    \begin{tcbclipinterior}
    \fill[blue!20!white] (frame.south west) rectangle ([xshift=20pt]frame.north west);
    \end{tcbclipinterior}},
  #3}

\usepackage{bm, physics, cancel}
\usepackage{thmtools, thm-restate}
\usepackage{caption}
\usepackage{array,graphicx,float, adjustbox}

\usepgfplotslibrary{groupplots}
\usepgfplotslibrary{patchplots}
\pgfplotsset{compat=1.16}
\newcommand{\R}{\mathbb{R}}

\newcommand{\cE}{\mathcal{E}}
\newcommand{\cN}{\mathcal{N}}

\newcommand{\cT}{\mathcal{T}}

\newcommand{\cL}{\mathcal{L}}
\newcommand{\cH}{\mathcal{H}}

\newcommand{\bD}{\mathbb{D}}

\newcommand{\bM}{\mathbb{M}}

\newcommand{\bT}{\mathbb{T}}

\newcommand{\bX}{\mathbb{X}}

\newcommand{\bZ}{\mathbb{Z}}

\newcommand{\x}{\times}

\definecolor{olive}{rgb}{0.6, 0.6, 0.2}
\definecolor{sand}{rgb}{0.8666666666666667, 0.8, 0.4666666666666667}
\definecolor{wine}{rgb}{0.5333333333333333, 0.13333333333333333, 0.3333333333333333}
\definecolor{deblue}{RGB}{11,132,147}
\definecolor{ocra}{RGB}{204, 119, 34}
\definecolor{depurple}{RGB}{131, 102, 135}
\definecolor{degrey}{RGB}{186, 172, 172}

\newcommand{\checkmarkf}{{\color{green!30!black}\checkmark}}
\newcommand{\xmarkf}{{\color{red!50!white}\xmark}}

\newtcolorbox{CatchyBox}[2][]{
    lower separated=false,
    colback=white!90!degrey!90!depurple,
    colframe=white, fonttitle=\bfseries,
    colbacktitle=white!60!degrey!90!depurple,
    coltitle=black,
    enhanced,
    attach boxed title to top left={xshift=.02\linewidth,yshift=-4mm},
    title=#2,#1}

\newtcbox{\mymath}[1][]{%
    nobeforeafter, math upper, tcbox raise base,
    enhanced, colframe=blue!30!black,
    colback=blue!30, boxrule=1pt,
    #1}

\title{Neural Hybrid Automata: Learning Dynamics with Multiple Modes and Stochastic Transitions}
\author{Michael Poli$^{1,3,}$\thanks{Equal contribution. Author order was decided by flipping a coin. $^1$KAIST $^2$The University of Tokyo  $^3$NAVER AI Lab $^4$ Naver Clova $^5$University of Toronto, Vector Institute $^6$Nvidia. Corresponding author: \textit{Michael Poli} ({$\tt poli\textunderscore m@kaist.ac.kr$})} , Stefano Massaroli$^{2,*}$, Luca Scimeca$^{3}$, Seong Joon Oh$^{3}$,\\ \textbf{Sanghyuk Chun$^{3,4}$, Atsushi Yamashita$^{2}$, Hajime Asama$^{2}$, Jinkyoo Park$^{1}$, Animesh Garg$^{5,6}$}}
\begin{document}
\maketitle
\doparttoc
\faketableofcontents

\begin{abstract}
Effective control and prediction of dynamical systems often require appropriate handling of continuous--time and discrete, event--triggered processes. Stochastic hybrid systems (SHSs), common across engineering domains, provide a formalism for dynamical systems subject to discrete, possibly stochastic, state jumps and multi--modal continuous--time flows. Despite the versatility and importance of SHSs across applications, a general procedure for the explicit learning of both discrete events and multi--mode continuous dynamics remains an open problem. This work introduces \textit{Neural Hybrid Automata} (NHAs), a recipe for learning SHS dynamics without \textit{a priori} knowledge on the number of modes and inter-modal transition dynamics. NHAs provide a systematic inference method based on normalizing flows, neural differential equations and self--supervision. We showcase NHAs on several tasks, including mode recovery and flow learning in systems with stochastic transitions, and end--to--end learning of hierarchical robot controllers.
\end{abstract}
\section{Introduction}
Behaviors emerging from the interaction of continuous and discrete--time dynamics in the presence of uncertainty are described through the language of \textit{stochastic hybrid systems} (SHSs). Such discrete events can bring along abrupt changes in the state, and in complex \textit{multi--mode} systems, may also cause a \textit{switch} between system \textit{modes}, and corresponding underlying continuous dynamics \cite{cassandras2018stochastic}. Communication networks \cite{hespanha2004stochastic,sun2004perturbation}, where changes in communication protocol can happen at certain levels of traffic congestion, and biological systems \cite{alur2001hybrid,hespanha2005stochastic,li2017review} are example domains where the SHS modeling paradigm has proven fruitful. 

Data--driven identification and learning of hybrid systems are known to be challenging due to the entanglement of continuous flows and discrete events \cite{lauer2019hybrid}; finding a generally applicable technique remains an open problem, particularly in the common scenarios where no \textit{a priori} knowledge on the number and type of \textit{system modes} is given. The aim of this work is to apply continuous neural models \cite{chen2018neural,massaroli2020dissecting,rackauckas2020universal} to the learning SHSs. We introduce a compact descriptive language for this task, decomposing the system into a set of core primitives. Prior work is integrated into the framework, highlighting in the process limiting assumptions and areas of further improvement.

To address the shortcomings of existing techniques, we introduce \textit{Neural Hybrid Automata} (NHA) as a general procedure designed to enable learning and simulation of SHSs from data. NHAs are comprised of three components: a {\color{deblue}dynamics} module, taking the form of an \textit{neural differential equation} (NDE) \cite{chen2018neural,massaroli2020dissecting,rackauckas2020universal} capable of approximating a different vector field for each mode, a {\color{olive}discrete latent state} tracking the internal mode of the target system, and an {\color{wine}event} module determining the time to next event. In particular, our approach does not require prior knowledge on the number of modes. The synergy among NHA components ensures a broader range of applicability compared to previous attempts, which in example do not directly tackle multi--mode hybrid systems \cite{jia2019neural,gwak2020neural,chen2020learning,zhong2021differentiable}. 
NHAs are shown to enable mode recovery and learning of systems with stochastic transitions, with additional applications in end--to--end learning of hierarchical robot controllers.
\section{Background}
We introduce required background on the formalism of \textit{stochastic hybrid systems} (SHSs) and event handling for their numerical simulation. We then provide further contextualization on previous approaches, introducing in the process a unified language for SHS learning tasks.
\subsection{Stochastic Hybrid Systems}\label{general_form}
A \textit{stochastic hybrid system} (SHS) \cite{hespanha2004stochastic,cassandras2007stochastic} is a right-continuous stochastic process $X_t$ taking values in $\bX\subseteq\mathbb{R}^{n_x}$ with a \textit{latent mode} process $Z_t$ conditioning the dynamics of $X_t$, where $t\geq 0$. $Z_t$ is another right-continuous stochastic process that takes values in a finite set $\mathbb{M}$ of size $m$. In this context, the set $\bM$ contains \textit{identifiers} of internal system \textit{modes}. An \textit{event} is defined as either a mode switch or a state discontinuity (a \textit{jump} in $X_t$), which can in some cases occur simultaneously. We refer to times at which events $z\rightarrow z'$ occur as random variables $t_k\in\cT$, with associated \textit{intensity functions}~\cite{daley2007introduction} 
\begin{equation*}
    \lambda_{z\rightarrow z'}(t|\cH_t)\geq 0.
\end{equation*}
where $\cH_t:=\{t_k\in\cT: t_k < t\}$ is the \textit{history} of event times. Intensity, as defined in the classical \textit{temporal point process} (TPP) sense, can be interpreted as the expected number of events $z\rightarrow z'$ within the time interval $[t,t+\dd t]$.
The dynamics for $X_t$ when $Z_t=z$ is defined by
\begin{equation*}
    {\color{gray!70!blue}\text{flow dynamics}}:~~~~~\dot x_t = F_z(t, x_t).~~~~~~~~~~~~~~~~~~~~~(z, t, x_t)\in\bM\times\bT\times\bX
\end{equation*}
When a jump event $z\rightarrow z'$ is triggered, $X_t$ can instantaneously jump according to 
\begin{equation*}
    {\color{gray!70!blue}\text{jump dynamics}}:~~~~x^+_t = \psi_{z\rightarrow z'}(t, x_t).~~~~~~~~~~~~~(z, z', t, x_t)\in\bM^{2}\times\bT\times\bX
\end{equation*}
Jump maps $\psi_{z\rightarrow z'}$ and intensities $\lambda_{z \rightarrow z'}$ describe the behavior during events $z\rightarrow z'$.
\subsection{Event Handling for Hybrid Systems}
Following \cite{shampine2000event}, to enable forward simulation of SHSs, a convenient mathematical representation of an \textit{event} is a function $g:\bT\x\bX\rightarrow\R$ which nullifies only at any \textit{event time} $t^*$, thus providing the differential equation integration algorithm with a \textit{termination} or \textit{restart} condition  i.e.
\begin{equation}\label{eq:event} 
    t^*\text{ is an event}~\Leftrightarrow~g(t^*, x_{t^*}) = 0.
\end{equation}
The particular form of $g$ induces a jump set on $\bD\subset\bT\times\bX, ~\bD:=\{t,x~:~g(t, x_t)=0\}$, and determines transitions from roots of $g$ to regions of the state--space $\bX$ where $g\neq0$. Notably, this construct enables utilization of root finding methods \cite{nocedal2006numerical} in a neighborhood of $t^*$ to accurately zero in on the event time.

The same simulation technique can be extended to the \textit{many} jump sets case typical of multi--mode systems, by equipping the condition function with identifiers $z,~z'$ ($g_{z\rightarrow z'}$) which induces jump sets $\bD_{z\rightarrow z'}$.
\vspace{-0.75em}
\paragraph{Simulating stochastic events}
While the event function approach appears to be limited to the deterministic setting, it also subsumes stochastic events whose aleatoric uncertainty is encoded by an intensity $\lambda(t | \cH_t)$ \cite{cassandras2018stochastic,chen2020learning}. Without loss of generality let us consider a single intensity function which is henceforth denoted as $\lambda^*_t:=\lambda(t | \cH_t)$. Recalling that the \textit{cumulative distribution function} (CDF) of inter--event times is $1 - \exp{-\int_{t_0}^{t^*}\lambda^*_t\dd t}$, standard \textit{inverse transform sampling} \cite{cassandras2018stochastic,rasmussen2011temporal} yields
\begin{equation}\label{eq:its}
   t^*: 0 = s - \log \int_{t_0}^{t^*}\lambda^*_t\dd t,~~s\sim {\tt Uniform}(0, 1)
\end{equation}
as a special case of \eqref{eq:event}. Approaches developed for learning TPPs, including in the context of Neural ODEs \cite{chen2018neural, jia2019neural, chen2020neural}, introduce a parametric formulation for the intensity $\lambda^*_\theta$ and optimize via direct TPP likelihood objectives. The integral is in general intractable, thus these methods require either a numerical approximation or the augmentation of additional states to compute it alongside the ODE. 

\subsection{Core primitives for SHS learning}\label{limitations}
At minimum, a learning model for SHSs necessitates several modules, each mirroring an element of the formulation in \ref{general_form}. More specifically:
\begin{itemize}[leftmargin=0.4in]
    \item[$i.$] {\color{deblue}Dynamics module}, to approximate continuous dynamics $F_z$ conditioned on mode $z$. 
    \item[$ii.$] {\color{olive}Discrete latent selector}, to identify at each event time the latent mode $z$ of the system.
    \item[$iii.$] {\color{wine}Event module}, to determine \textit{when} events happen, and how state $x$ and latent state $z$ are updated after the transition.
\end{itemize}
Prior work considers specific instantiations of SHSs, leading to simplifying choices for each module defined above. In example, switching systems without jumps, where $iii.$ does not require jumps \cite{lauer2011continuous}, single mode systems, where $ii.$ is not required and $iii.$ does not need to determine latent mode transitions \cite{jia2019neural,chen2020learning,zhong2021differentiable}, systems with known dynamics, where $i.$ is not trained \cite{chen2020learning}, or systems with only deterministic events \cite{lauer2019hybrid}. We note some of these works suffer from more than a single of these limitations, including additional ones consequence of specific model choices. Direct parametrization of the intensity, while a reasonable choice for single mode systems, requires state augmentation scaling in the worst case as ${m \choose 2}$ \cite{west2001introduction} for a SHS with $m$ modes. More importantly, training parameters $\theta$ for such direct approaches is affected by the accuracy of the numerical method employed for the solution of the integral in \eqref{eq:its}. 

In this work, we introduce a modelling framework SHSs that does not rely on any of the simplifying assumption on $i.$, $ii.$ and $iii.$ outlined above. Furthermore, our method scales as $O(m)$ in the number of SHS modes.
\begin{figure}[t]
    \centering
    \includegraphics[width=.9\linewidth]{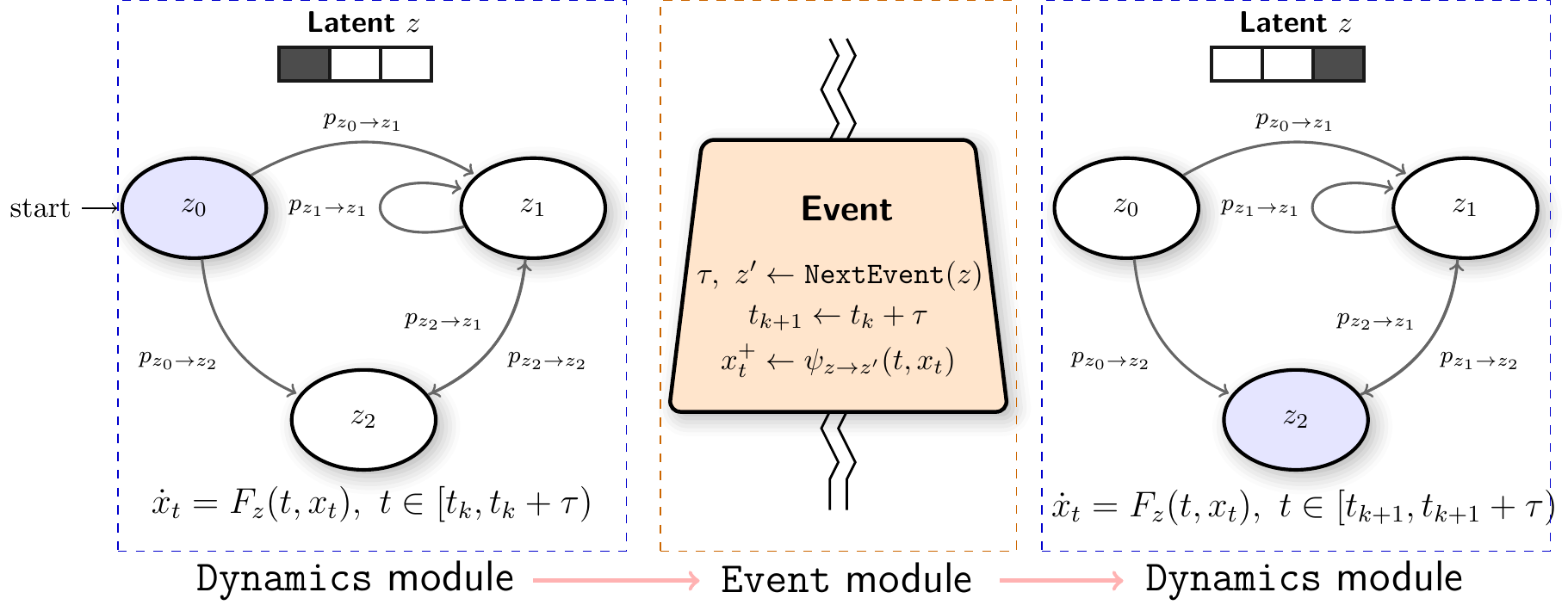}
    \vspace{-2mm}
    \caption{\footnotesize  Schematic of a Neural Hybrid Automata (NHA). The mode--conditioned Neural ODE $F_z$ drives the system forward until an event time $t_{k+1}$ determined by a previous call to the event module. Then, the event module determines time to next event and corresponding mode target $z'$ through sampling from normalizing flow $p_{z\rightarrow z'}$ approximating densities of interevent times. A jump function is then applied to the state $x$, and simulation continues with flow $F_{z'}$.}
    \label{fig:catchy}
    \vspace{-10pt}
\end{figure}

\section{Neural Hybrid Automata}
We introduce \textit{Neural Hybrid Automata} (NHA), a model for learning of SHSs. A NHA is comprised of a {\color{deblue}dynamics module}, a {\color{olive}discrete latent state} and an {\color{wine}event predictor}. A general overview of a NHA is depicted in Figure \ref{fig:catchy}. We start with a description of each module and their interconnections, followed by a step--by--step procedure for NHA training.
\vspace{-0.75em}
\paragraph{Event module}
Intensity--free parametrizations for stochastic mode transitions allows NHAs to sample next event times without solving integrals $\int \lambda_t^* \dd t$ for all target modes $z'$ reachable from current $z$. From event $k$ at time $t_k$, NHAs determine next event times $t_{k+1}$ through a conditional normalizing flow modeling, for each possible pair of $(z, z')$, the density of corresponding \textit{inter--event} times $\tau^{k}_{z\rightarrow z'} = t_{k+1} - t_k, t_k\in\cT_{z\rightarrow z'}$. 
Let the intensity be a simple \textit{timer} i.e $\dot\lambda = 1$. Further, let $p_{z\rightarrow z'}$ be the parametrized conditional density obtained by the normalizing flow and let $T(z, z', t_k)$ be a collection of conditional samples from $p_{z\rightarrow z'}$ (one for each pair $z,~z'$), i.e. 
%
$$
    T(z, z', t_k) = \Big\{\tau^k_{z\rightarrow z'}\sim p_{z\rightarrow z'}(\tau| \cH_{t_k})\Big\}_{z,z'\in\bZ}.
$$
Using \eqref{eq:its}, we can thus sample an event given the current mode $z$ and the previous event time $t_k$ as:
\begin{equation}\label{eq:itsv2}
\begin{aligned}
      t_{k+1} = t_k + \min_{z'\in\bZ}T(z, z', t_k).
\end{aligned}
\end{equation}
Note that the next mode $z'$ after the event is simultaneously obtained as $z'=\arg\min T(z, z', t_k)$.
Sampling strategy \eqref{eq:itsv2}, differently from \eqref{eq:its}, relies on the normalizing flow to explicitly model the density rather than defining it implicitly through $\int \lambda^*_t \dd t$. When event time $t_k$ is reached, a parametrized jump map conditioned on $(z,z')$ is applied to the state $x^+=\psi_{z\rightarrow z'}(t_k, x)$. Normalizing flow $p_{z\rightarrow z'}$ and jump map $\psi_{z\rightarrow z'}$ together define the full event module of an NHA.

It should be noted that \eqref{eq:itsv2} always samples the \textit{quickest--to--occur} event from the normalizing flow, which implies that no other event occurs between $t_k$ and $t_{k+1}$. While the history can be compressed into a fixed--length vector following \cite{shchur2019intensity} through application of sequence models e,g. RNNs, we note that for hybrid systems equipped with deterministic events, providing $(x_{t_k}, t_k)$ as conditioning inputs for $p_{z\rightarrow z'}(\tau)$ is sufficient since ODE solutions with deterministic transitions are uniquely determined by the initial condition. Finally, deterministic events are a special case of stochastic events \cite{hespanha2004stochastic} that can be well--represented with a Dirac $\delta$ function, of which the normalizing flow learns a smooth approximation with continuous support.

\vspace{-0.75em}
\paragraph{Dynamics module}
To enable approximation of different mode--dependent vector fields, we parametrize the \textit{flow map} $F_z(t, x_t)$ of a SHS as a \textit{data--controlled neural ordinary differential equation} (Neural ODE) \cite{massaroli2020dissecting} with parameters $\omega$, driven between each pair of event times $t_{k}, t_{k+1}$ by discrete latent mode $z$
\begin{equation}\label{eq:syss1}
    \begin{aligned}
        \dot x_t &= F_z(t, x_t, \omega)~~~&t&\in[t_{k}, t_{k+1})
    \end{aligned}
\end{equation}
Finiteness of admissible values in the latent mode state i.e. $m$ ensures $F$ is capable of approximating a finite number of different vector fields, one for each mode. In particular, we consider one--hot representations for latents $z\in\R^{m}$. In batched data settings, \eqref{eq:syss1} can be integrated in parallel across $n_b$ batches of initial conditions $x_{t_k}\in\R^{n_b\times n_x}$ with different modes, provided the latent is also batched $z\in\R^{n_b\times m}$.
The combination of a given dynamics and event module, applied in turn as depicted in Figure \ref{fig:catchy}, enables simulation of trajectories of a SHS. We now describe their training procedure.
\begin{figure}[t]
    \centering
    \includegraphics[width=0.95\textwidth]{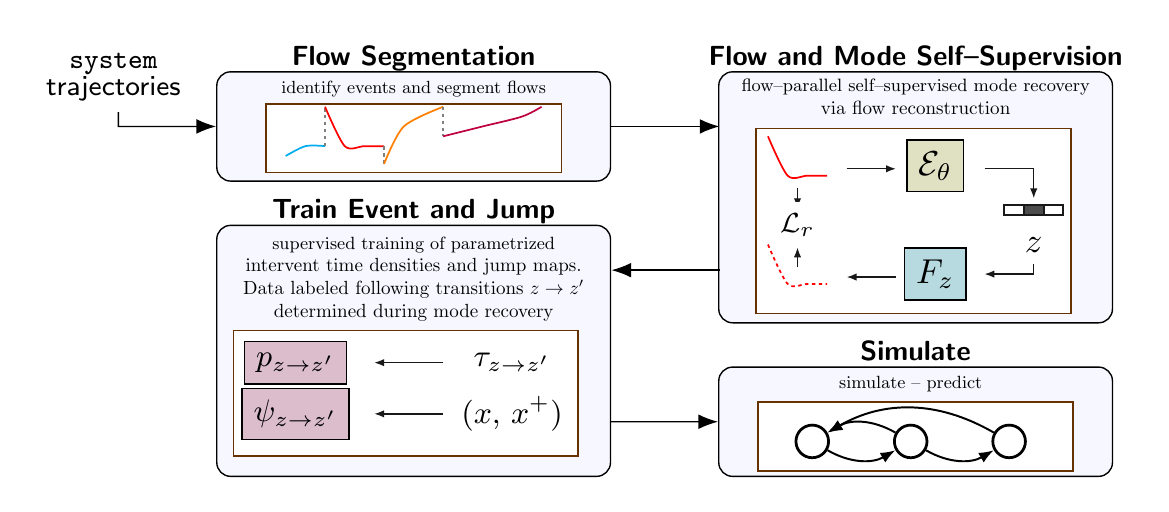}
    \vspace{-4mm}
    \caption{\footnotesize NHA training blueprint. Segmenting the trajectories enables self--supervised mode recovery via trajectory reconstruction. The recovered mode labels are then used for NHA event module supervised training.}
    \vspace{-6mm}
    \label{fig:recipe}
\end{figure}

\section{Neural Hybrid Automata Module Training}
Here, we detail the training procedure for each NHA component. Our only assumption is availability to a \textit{trajectory segmentation} routine tasked with separating the trajectories, or flows, into a collection of \textit{subtrajectories} $X_i$ of potentially of different length, each produced by the system in a different mode. The routine can be as simple as detection of discontinuities in the solution by inspecting \textit{finite--differences} of observations across timestamps \cite{massaroli2020identification}, or involve additional steps such as change--point detection \cite{aminikhanghahi2017survey}.
Providing exact event times to NHAs is not required; the segmentation routine need only partition the full dataset in $n$ disjoint sets $X_i$ s.t. $\bigcup_iX_i=X$ and $\bigcap_iX_i=\emptyset$. In addition, no knowledge of the number of modes, or topology of transitions between modes is made available to NHAs, as these are rarely available in practice.
\vspace{-0.75em}
\paragraph{Self--supervised mode recovery}

The first stage of learning an NHA is designed to approximate the continuous dynamics under each SHS mode while simultaneously identifying modes $z$. We achieve this by framing subtrajectory reconstruction as a \textit{pretext} task for mode recovery, via a reconstruction objective $\cL_r = \frac{1}{n}\sum_{i=0}^{n}\ell_r(X_i, \hat{X}_i)$, being $\hat{X}_i$ subtrajectories reconstructed by the {\color{deblue} flow decoder} $F_z$ via the model 
\begin{equation}\label{eq:syss}
    \begin{aligned}
        z &\sim \cE(X, \theta)~~~&t&=t_{k}\\
        \dot x &= F_z(t, x_t, \omega)~~~&t&\in[t_{k}, t_{k+1}).
    \end{aligned}
\end{equation}
Here, a {\color{olive}latent \textit{encoder}} $\cE$ with parameters $\theta$ is tasked with extracting a latent mode state $z\in\bM$ to steer the decoder $F_z$ towards a more accurate reconstruction. Representation limitations of Neural ODEs \cite{dupont2019augmented,massaroli2020dissecting} ensure that to fit the above objective the the {\color{olive}encoder} $\cE$ has to cluster the trajectories to enable the data--control decoder to represent different vector fields for each system mode.
Finiteness of admissible values in the latent state is enforced by defining $z$ as one--hot encoded sample from a parametrized categorical distribution. Backpropagating through the sampling procedure is performed via straight--through gradients \cite{bengio2013estimating}. System \eqref{eq:syss} can be regarded as an ODE trajectory autoencoder with a categorical bottleneck.

Finally, we note that trajectory segmentation serves multiple purposes during NHA training. Forward integration is significantly sped up since the ODE solves can now be parallelized across subtrajectories $X_i$ as independent samples of a batch of data, avoiding a sequential solve on full SHS trajectories. The speedups can be dramatic for multi--mode SHSs\footnote{While speedups are dependent on full trajectory and average subtrajectory lengths, in our experiment we observe at least an order of magnitude (more than $20$x) in wall--clock speedups for a single training iteration.}, the focus of this work, where data trajectories may need to be longer to sufficiently \textit{explore} different modes.
\vspace{-0.75em}
\paragraph{Event and jump supervision}
In addition to the learning of mode dynamics, self--supervised mode recovery objectives provides direct supervision for normalizing flows $p_{z \rightarrow z'}$ and jump maps $\psi_{z \rightarrow z'}$. More specifically, we collect times $\tau^k_{z\rightarrow z'}$ and jump state pairs $(x, x^+)$ for each pair of modes $(z, z')$ corresponding to a transition between pairs of subtrajectories clustered as $z$ (first) and $z'$ (second) by the encoder $\cE$. We then train the jump maps $\psi_{z\rightarrow z'}$ to approximate $x\mapsto x^+$, and the mode conditional normalizing flow to approximate the density $p_{z\rightarrow z'}(\tau)$.

\vspace{-0.75em}
\paragraph{Gradient pathologies in joint learning of flows and events}
\begin{wrapfigure}[10]{r}{0.5\textwidth}
    \centering
    \vspace{-9mm}
    \includegraphics[width=1\linewidth]{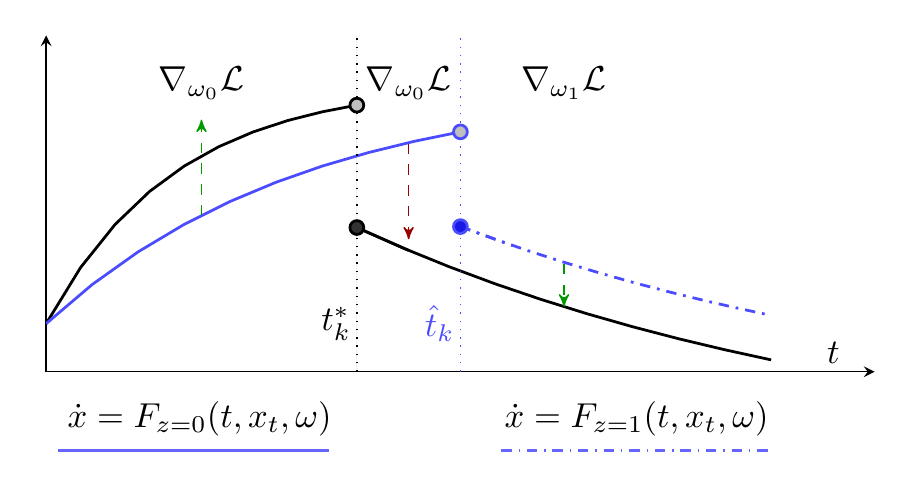}
    \vspace{-9mm}
    \caption{Conflicting gradient information in an idealized $2$--mode hybrid system due to overestimation of event time.}
    \label{fig:gradpat}
\end{wrapfigure}

When attempting simultaneous learning on the full trajectory, the parameters of flow $F_z$ can be subjected to wrong gradients from reconstruction objectives, arising from overreliance or underreliance of the flow model on certain modes. This phenomenon can bias training, and provides strong motivation behind our segmentation--first approach, since each subtrajectory $X_i$ is associated only to a single mode\footnote{Due to inaccuracies or noise in the segmentation algorithm, these partitions might not be perfectly separated into different modes. We experimentally investigate these effects on NHA training in Appendix B.}.

A visualization is provided in Figure \ref{fig:gradpat}, through an idealized learning task of a two--mode system. Overreliance of the flow model on mode $z=0$, due to overestimation of event time $t_k$, leads to a decomposition of gradients $\nabla_{w_0}\mathcal{L}_r$; in {\color{green!70!black}green}, gradients pushing the trajectory closer to the solution, in {\color{red!70!black}red}, incorrect gradients pushing the mode $0$ trajectory further away from the ground--truth and closer to a solution belonging to a different mode. Appendix A further develops theoretical considerations on the nature of these gradients.

\section{Results and Discussion}
We evaluate \textit{Neural Hybrid Automata} (NHA) through extensive experiments, with a focus on investigating the performance and robustness of each NHA module. A summary of experiments, objectives and ablations is provided here for clarity:
\begin{itemize}[leftmargin=0.2in]
    \item \textbf{Reno TCP}: we carry out a quantitative evaluation on quality of learned flows (mean squared error) and quality of mode clusters recovered during self--supervision (v--measure). We also verify the robustness of NHAs to overclustering and amount of data required for event module training.
    \item \textbf{Mode mixing in switching systems}: we highlight and varify robustness against \textit{mode mixing}, a phenomenon occurring during learning of multi--mode systems through alternative \textit{soft} parametrization of latent $z$, such as through {\tt softmax} instead of {\tt categorical} samples.
    \item \textbf{Behavioral control of wheeled robots}: NHAs enable task--based behavioral control. We investigate a point--to--point navigation task where a higher level \textit{reinforcement learning} (RL) planner determines mode switching for a lower--level optimal controller. 
\end{itemize}

\begin{figure}[t]
    \centering
    \includegraphics[width=\textwidth]{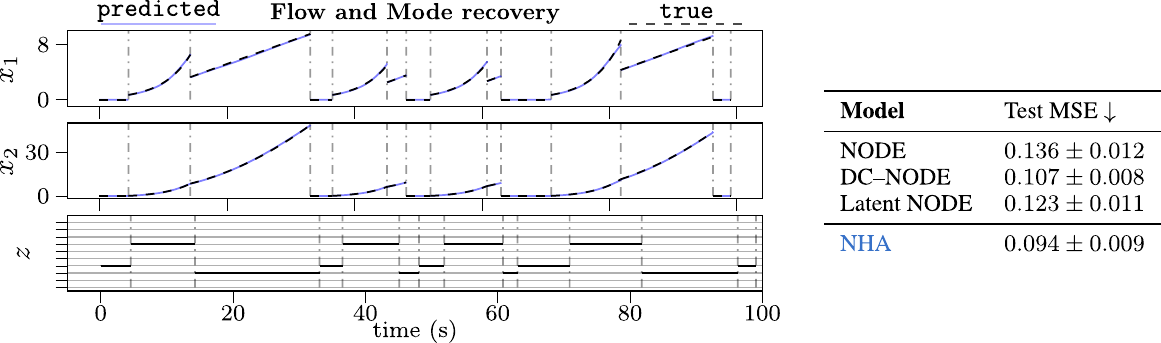}
        \vspace{-2mm}
    \caption{\footnotesize \textbf{[Left]} Reconstruction of system trajectories through NHA vector field decoders $F_z$ and corresponding modes $z$ encoded by $\cE$ for Reno {\tt TCP}. Although the encoder is initialized with more modes ($10$) than there are in the underlying system ($3$), mode clustering is sparse and accurate. \textbf{[Right]} Flow reconstruction test MSE for different classes of decoders. NHA decoders ($10$ modes) can reconstruct the flows as well as other NODE baselines, with the added benefit of being able to recover mode labels during training.}
    \label{fig:recon}
    \vspace{-5mm}
\end{figure}
\subsection{System with Stochastic Transitions}
\begin{wrapfigure}[13]{r}{0.45\textwidth}
    \centering
    \vspace{-16mm}
    \includegraphics[width=\linewidth]{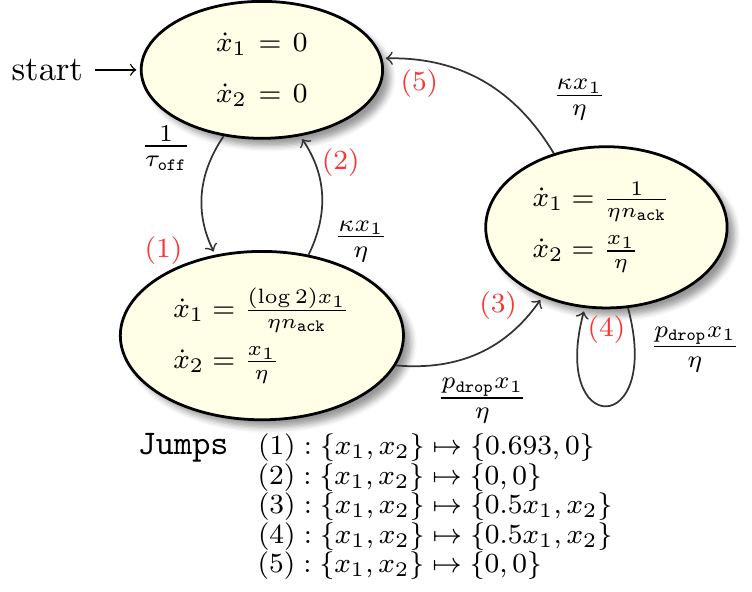}
    \vspace{-6mm}
    \caption{\footnotesize Automata representation of TCP Reno, where $\eta,~p_{\tt drop},~\kappa > 0$ and we set $n_{ack}=2$. On each edge is the corresponding intensity $\lambda_{z\rightarrow z'}$.}
    \label{fig:tcpautomata}
\end{wrapfigure}
\begin{figure}[t]
    \centering
    \includegraphics[width=1\linewidth]{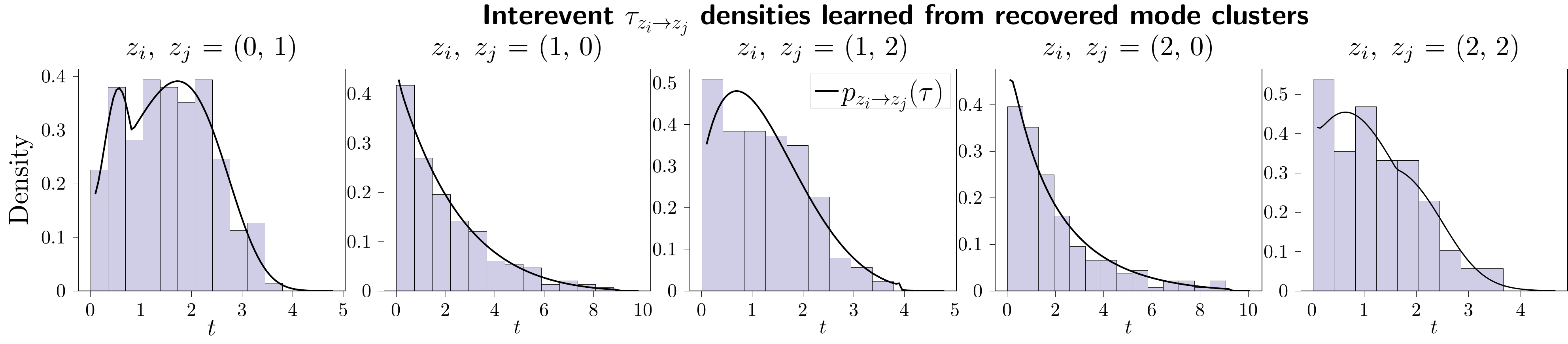}
    \vspace{-6mm}
    \caption{\footnotesize Learned densities (black) for intervent times $\tau_{z_i \rightarrow z_j}$. The normalizing flows are trained on times recovered during the mode recovery stage, by clustering $\tau_{z_i \rightarrow z_j}$ according to the encoder mode labeling. The histogram depicts the ground--truth empirical distribution for each class of event.}
    \label{fig:qualp}
    \vspace{-10pt}
\end{figure}

We apply NHAs to a dataset of internal state trajectories of a network transmission controller (TCP), the Reno TCP scheme \cite{hespanha2004stochastic}. The system has two states, five stochastic transitions and three modes as shown through an automata representation in Figure \ref{fig:tcpautomata}. Here, we qualitatively validate the performance of dynamics and event modules of NHAs. We simulate $40$ trajectories of the system, each $200$ seconds long, and segment them. No a priori knowledge on the mode of each subtrajectory is provided to the model. We perform self--supervised mode recovery to train $F_z$ and $\cE$, in the process labeling the subtrajectories, then train event module normalizing flows and jump functions with the mode labels obtained. Training and evaluation are performed using a $5$--fold cross--validation strategy, with a final test fold of $15$. More details on the system, architectures and data generation are reported in the Appendix B.
\vspace{-0.75em}
\paragraph{Mode recovery results}
First, we perform self--supervised mode recovery and verify (i) whether the mode conditioned NHA decoder $F_z$ offers test--time TCP trajectory reconstruction of equal or better quality than other Neural ODE variants, and (ii) quality of the mode label clusters assigned by the NHA encoder and robustness to different mumber of latent modes $m$. We measure (i) via test \textit{mean squared error} (MSE) on reconstructed trajectories, and (ii) via \textit{v--measure} \cite{rosenberg2007v}, a metric taking values in $[0,1]$, computed as the harmonic mean between cluster \textit{completeness} and \textit{homogeneity}. A v--measure of $1$ indicates perfect clustering. As baselines, we collect for (i) the 
performance of $3$ \textit{Neural ODE }(NODE) variants, a zero--augmented NODE, a \textit{data--controlled NODE} (DC--NODE) \cite{massaroli2020dissecting} where the latent $z$ is the output of a multi--layer encoder, and a Latent NODE where $z$ is sampled via reparametrization of a Normal \cite{chen2018neural}. 
We also provide baseline performance of a series of popular clustering algorithms tasked to cluster the subtrajectories: k--means++ \cite{arthur2006k}, hierarchical \cite{murtagh2012algorithms} and DBSCAN \cite{birant2007st}. Figure \ref{fig:recon} provides qualitative and quantitative results for stage (i). As established by \ref{tab:res1}, NHA mode recovery outperforms all baselines by a wide margin, with v--measure values close to $1$. Surprisingly, we observe providing the NHA encoder with a larger number of latent modes than the $3$ of the system improves clustering results.
Additional details on data pre--processing, metrics and baseline design and tuning are provided in Appendix B.
\begin{wraptable}[22]{r}{0.6\linewidth}
    \footnotesize
    \centering
    \begin{tabular}{llllll} \toprule
         &   & {\tt v}--measure $\uparrow$ & \\ \midrule
        \textbf{Model} & $m=3$ & $m=5$ & $m=10$ \\ \midrule
        k--means{\tt $++$} & $0.20 \pm 0.02$ & $0.24 \pm 0.02$ & $0.30 \pm 0.06$  \\
        hierarchical & $0.23 \pm 0.01$ & $0.24 \pm 0.01$ & $0.31 \pm 0.06$ \\
        DBSCAN & $0.66 \pm 0.02$ & $0.68 \pm 0.02$ & $0.69 \pm 0.01$ \\ \midrule
        NHA & $\mathbf{0.86} \pm 0.02$ & $\mathbf{0.91} \pm 0.02$ & $\mathbf{0.96} \pm 0.03$ \\ 
    \end{tabular}
    \caption{\footnotesize Quality of recovered mode clusters from NHA self--supervised training and baseline clustering algorithms in the TCP task. Hyperparameter $m$ is the number of clusters provided to each algorithm. For DBSCAN, values of $m\in[3,~5,~10]$ map instead to its primary parameter $\epsilon\in[0.1,~0.5,~1]$ \cite{birant2007st}.}
    \label{tab:res1}
    \vspace{2mm}
    \begin{tabular}{llllll}\toprule
        \footnotesize
        \textbf{Model} & \textbf{Metric} & $n=1$ & $n=3$ & $n=5$ & $n=10$ \\ \midrule
        $p_{z\rightarrow z'}$ & NLL $\downarrow$ & $2.761$ & $2.375$ & $2.362$ & $2.313$ \\
        $\psi_{z\rightarrow z'}$ & MSE ($10^{-3}$) $\downarrow$ & $1.435$ & $0.018$ & $0.009$ & $0.003$ \\
    \end{tabular}
    \caption{\footnotesize Quality of fit for event module components, normalizing flows $p$ and jump maps $\psi$. Training performed with supervising mode labels from $n$ trajectories of TCP. We report test MSE and \textit{negative log--likelihood} (NLL) estimated from a base normalizing flow model trained on ground--truth data from $n=500$.}
    \label{tab:res}
\end{wraptable}
\vspace{0.75em}
\paragraph{Event module results}
Next, we leverage the mode labels recovered as supervision for the event module of an NHA. In all cases, we train three--layer MLPs as jump maps and two--layer spline flows \cite{durkan2019neural} as normalizing flows. Figure \ref{fig:qualp} visualizes the learned densities for each stochastic transition for the standard training regime of $n=5$ trajectories. We also perform an ablative study on the quantity of data required to extract sufficient supervision signal for both components of the event module. The results are included in Table \ref{tab:res1}. We find that a single trajectory is sufficient, with relative performance gains quickly dropping off after $n=3$.

\subsection{Deterministic Switching System}
We investigate \textit{mode mixing} in a three--mode switching linear system (SLS) \cite{chen2020learning}. The deterministic nature of mode transitions, as well as the absence of state jumps enables direct training of NHAs on data--trajectories without prior segmentation. This allows us to perform an ablative study on \textit{mode mixing}, a phenomenon arising from latent modes $z$ produced by the encoder $\cE$ via \textit{soft} alternatives to categorical samples, such as by using {\tt softmax} activations. 

\begin{figure}[t]
\centering
\begin{minipage}[c]{0.7\textwidth}
    \centering
     \includegraphics[width=\linewidth]{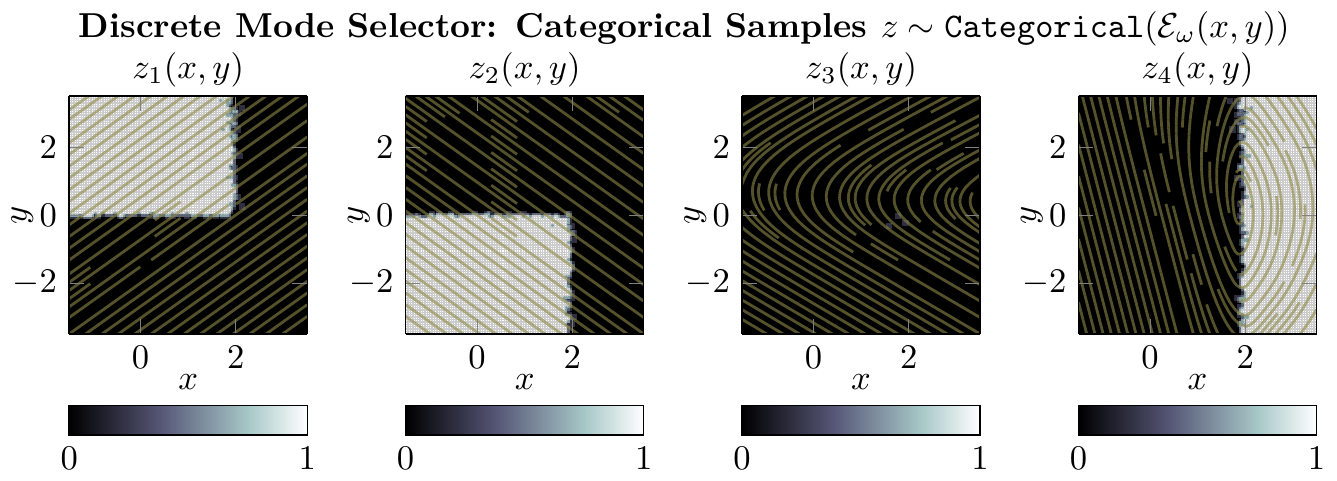}
    \includegraphics[width=\linewidth]{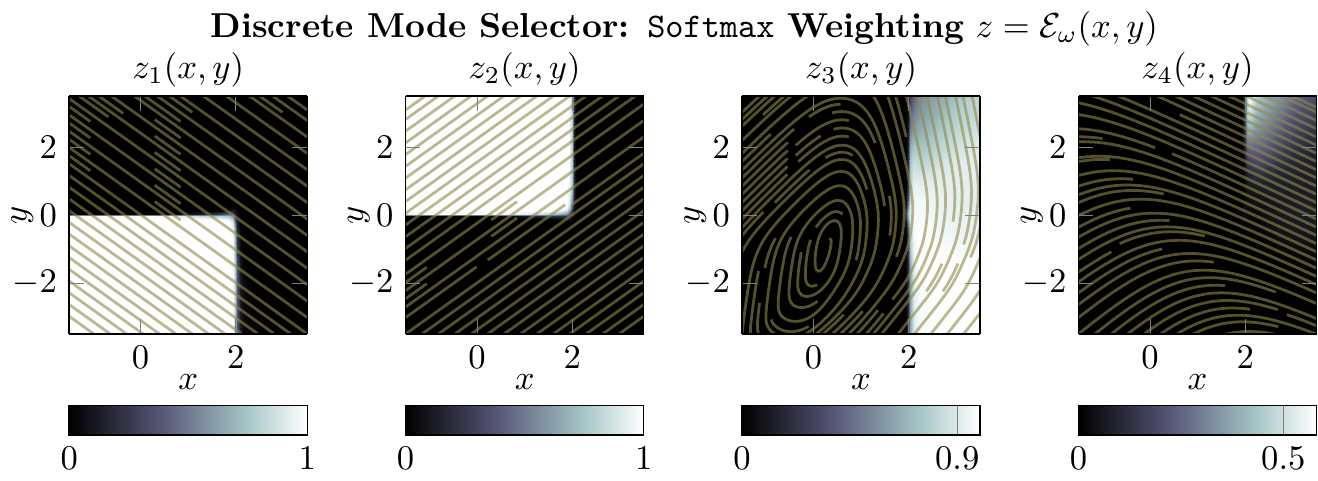}
\end{minipage}
~
\begin{minipage}[c]{0.28\textwidth}
    \centering
    \caption{Reconstructed conditional vector fields $F_z$ and corresponding mode classification boundaries in the state--space of the LSS. \textbf{[Above]} Categorical NHA encoder. \textbf{[Below]} Mode classification performed by "soft" encoder $\cE$ capped with a {\tt softmax} activation.}
    \label{fig:mode_mixing}
\end{minipage}
 
\end{figure}

\vspace{-0.75em}
\paragraph{Mode mixing and overclustering}
We train NHAs on reconstruction of SLS trajectories. Each encoder is provided, at initialization, one additional latent mode over the three of the system. The conditioned flow $F_z$ is constructed with three--layer MLPs. Rather than segmenting the data, we repeat sampling for $z$ at each integration step. Figure \ref{fig:mode_mixing} shows the state space switching boundaries and
mode vector fields learned by an NHA and a baseline producing $z$ via {\tt softmax} rather than as categorical samples. 
The additional freedom provided by softmax latents $z\in\mathbb{R}^4_+,~\sum z_i = 1$ allows fitting the trajectories by nonlinearly mixing different vector fields to approximate a single one. Instead, categorical samples cannot mix the vector fields; this ensures that the learned clustering is either sparse as shown in the Fig.\ref{fig:mode_mixing}, or latent values dedicated to the approximation of the same underlying mode dynamics are forced to learn the same vector field. Appendix B contains a visualization and analysis for this second case.

In general, categorical sampling is effective when recovery of system modes is a model objective, whereas {\tt softmax} or other soft relaxations can be viable if only black--box fitting of data is desired.

\begin{figure}[b]%
    \vspace{-10pt}
    \centering
    \subfloat{%
    \label{fig:ex3-a}%
    \includegraphics[scale=1]{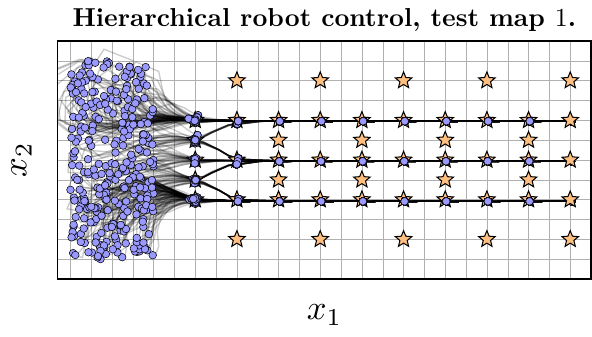}}
    \hspace{1pt}%
    \subfloat{%
    \label{fig:ex3-b}%
    \includegraphics[scale=1]{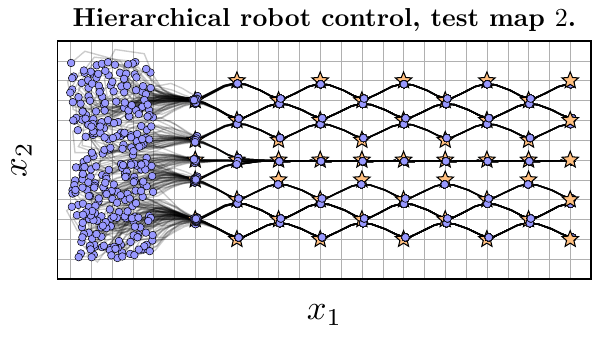}}\\
    \hspace{1pt}\\%
    \vspace{-9mm}%
    \subfloat{%
    \label{fig:ex3-d}%
    \includegraphics[scale=1]{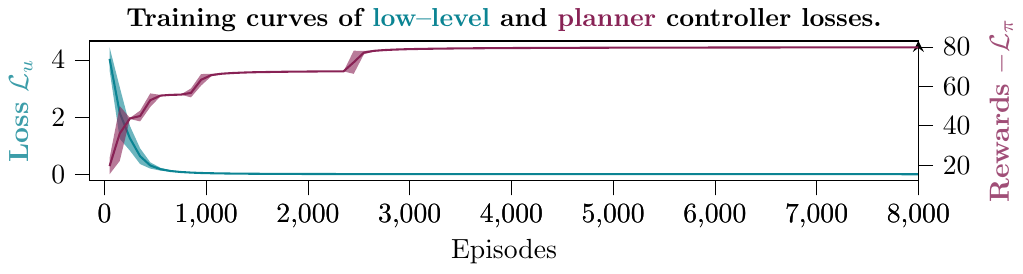}}%
    \vspace{-3mm}
    \caption{\footnotesize \textbf{[Above}]: Test--time learned navigation of swarms of differential drive robots. The robots are initialized at random locations and orientations. \textbf{[Below}]: Training loss curves of low--level controller $u$ and planner $\pi$ across episodes.}%
    \label{fig:ex3}%
\end{figure}

\subsection{End--To--End Learning of Hierarchical Switching Controllers for Dynamical Systems}
Beyond SHS identification, the NHA framework enables learning of task--based hierarchical controllers comprising a low--level controller $u_z := u(t, x_t, z)$ dependent on the discrete mode $z$ provided by a higher--level policy $\pi$. Each NHA module is adapted as:
\begin{equation}
    \begin{aligned}
    {\color{deblue}\text{dynamics module:}}~~\dot x_t &= F(t, x_t, u_{z})\dd t ~~&&\text{low--level controlled system}\\
    {\color{wine}\text{event module:}}~~z' &\leftarrow \pi(t, x_t, z) ~~&&\text{high--level planner}
    \end{aligned}
\end{equation}
Within this context, {\color{olive}latent state} $z$ can be regarded as a system \textit{set--point} (e.g. a desired value of state $x$) determined by the planning policy $\pi$ to achieve a certain task, which the low--level controller has then to carry out. Both $\pi$ and $u_z$ are parametrized by neural networks, and the training is done end--to--end. The obtained hierarchical control scheme is sample efficient, since system dynamics available a priori are included in $F$. From the perspective of $\pi$, however, the dynamics are \textit{disentangled} from the planning objective. Indeed, the higher level policy need only learn how to set and switch between objectives ($z\rightarrow z'$), and not how to control the system to reach them. For this reason we train the model using a loss function partitioned in two terms as $\cL = \cL_u + \cL_{\pi}$.

\vspace{-0.75em}
\paragraph{Results}
We consider learning controllers for navigation of two-wheeled differential drive robots \cite{malu2014kinematics}. The higher--level model--free policy $\pi$ is here trained via REINFORCE gradients \cite{sutton2018reinforcement,peters2006policy} to select the nearest resource target every $5$ seconds. We set 5 different resources within a map, and train all robots to: one, select the nearest resource target; two, drive the wheels to reach the target.  The training of both controllers is carried out concurrently, where target (or \textit{mode}) selection is performed by $\pi$, whereas the behavioural controller $u_z$ has to reach the target chosen by $\pi$ via low--level steering control inputs. As shown in Figure \ref{fig:ex3}, convergence of both control policies occurs after around $4000$ episodes of training. Figure \ref{fig:ex3} also visualizes the resulting navigation behavior at test--time on two new resource layouts, where we alternate between two different sets of targets.  

\subsection{Generalizable Insights and Empirical Observations}
The task of learning SHSs involves several moving components. Ablative experiments have been performed to address specific questions on the robustness of NHAs. We detail some {\color{orange!80!white}empirical heuristics} observed to improve performance, and report {\color{purple!80!white}areas of further improvement}. 
\vspace{-0.75em}
\paragraph{\textcolor{orange!80!white}{\textbullet}~Overclustering stabilizes training}
We empirically observe providing NHA with more latent modes than the system stabilizes training. We conjecture the additional choice allows the model to use different modes during exploration without always conflicting with other already "assigned" modes, phenomenon which is more frequent, in example, when the number of modes exactly matches that of the system. In these cases, dropout in the encoder $\cE$ appears to improve performance.
\vspace{-0.75em}
\paragraph{\textcolor{purple!80!white}{\textbullet}~Noisy segmentation of trajectories}
We investigate, for the TCP experiment, robustness of mode recovery to incorrect segmentation (Appendix B) and number of NHA latent modes $m$ (Table \ref{tab:res1}). Extending NHAs to include a finetuning step for trajectory segmentation, in example leveraging ideas from \cite{chen2020learning} might improve robustness of the segmentation routine and thus the overall approach.

\section{Related Work}

\vspace{-0.75em}
\paragraph{Hybrid system identification and Markov models}
Hybrid system identification is a relatively recent development in dynamical system theory \cite{lauer2019hybrid}. A majority of existing literature focuses on (linear) \textit{piecewise affine systems} (PWA) \cite{paoletti2007identification,westra2011identification}. \cite{ly2012learning} proposes a clustered symbolic regression algorithm for learning input--output maps rather than dynamics. Existing approaches involving continuous optimization \cite{lauer2011continuous} do not consider event stochasticity and mode recovery. Identification of SHSs is an even smaller field, with limited success outside specific cases \cite{cassandras2007stochastic}. 
\vspace{-0.75em}
\paragraph{Continuous--depth and contact models}
Neural differential equations and continuous--depth models, initially concerned with unimodal systems \cite{chen2018neural,massaroli2020dissecting}, have seen preliminary application to the learning of temporal point processes \cite{rubanova2019latent,jia2019neural}. Although some of these works tackle stochastic events and marked point processes, multimodality and explicit learning of the flows is not considered. \cite{gwak2020neural} examine \textit{interventions} as events, and develop a continuous architecture for modeling the lasting effect of a given intervention on the dynamics. Differentiable contact approaches \cite{zhong2021differentiable,hochlehnert2021learning} introduce physics--compatible models designed to recover deterministic hybrid dynamics of mechanical systems from data. \cite{chen2020learning} develops, through implicit differentiation, a method for direct optimization of event times. Although Neural Event ODEs do not directly address multimodality, a potential synergy between the approach of \cite{chen2020learning} and NHAs could preserve the advantage of our flow--parallel mode recovery, namely integration speed and sidestepping of gradient pathologies outlined in Section IV. Section \ref{limitations} provides a summary of limitations for these existing methods.

\section{Conclusion}
Hybrid systems represent a versatile and general class of systems, with applications across engineering disciplines \cite{cassandras2007stochastic}. In this work, we investigate challenges related to the learning of SHSs from data and introduce \textit{Neural Hybrid Automata} (NHA), a step--by--step method leveraging neural differential equations, density estimation and self--supervised system mode recovery. NHAs are shown to be effective in various settings, including flow and event learning in systems with stochastic transitions.

\clearpage
\newpage
\printbibliography
\clearpage
\newpage
\rule[0pt]{\columnwidth}{3pt}
\begin{center}
    \huge{\bf{Neural Hybrid Automata} \\
    \emph{Supplementary Material}}
\end{center}
\vspace*{3mm}
\rule[0pt]{\columnwidth}{1pt}
\vspace*{-.5in}

\appendix
\addcontentsline{toc}{section}{}
\part{}
\parttoc
%
\section{Additional Discussion and Theory}
\subsection{Neural Hybrid Automata: Modules and Hyperparameters}
We provide a notation and summary table for \textit{Neural Hybrid Automata} (NHA). The table serves as a quick reference for the core concepts introduced in the main text. 
\begin{itemize}[leftmargin=0.4in]
    \item[$1.$] {\color{deblue}Dynamics}: tasked with approximating continuous dynamics of each mode by conditioning a Neural ODE on mode $z$.  
    \item[$2.$] {\color{olive}Mode Encoder}: only used during self--supervised mode recovery. Labels every subjtrajectory $X_i$ with a mode $z$ to ensure mode--conditioned decoder $F_z$ can reconstruct it despite Neural ODE representation limitations (uniqueness of solutions given an initial condition). 
    \item[$3.$] {\color{wine}Event Module}: determines during simulation (i) when events happen, and what types of events i.e. mode transitions $(z\rightarrow z')$ through $p_{z\rightarrow z'}$, (ii) what happens during such events i.e. jumps on the state via $\psi_{z\rightarrow z'}$. Normalizing flow $p_{z\rightarrow z'}$ is trained to approximate densities $p(\tau_{z\rightarrow z'} | \cH)$.
\end{itemize}
The only NHA hyperparameter beyond module architectural choices is $m$, or number of latent modes provided to the model at initialization. Performance effects of changing $m$ have been explored in Section 5.2 and Appendix B.2. Appendix B.2 further provides analyzes potential techniques to prune additional modes. 
\begin{CatchyBox}{Neural Hybrid Automata Modules}
    \begin{minipage}[h]{0.50\linewidth}
            {\color{deblue}Dynamics}: ~~~$\dot x = F_z(t, x_t, \omega)~~~t\in[t_{k}, t_{k+1})$
            \\
            
            {\color{olive}Mode Encoder}: ~~~$z \sim \cE(X, \theta)~~~t=t_{k}$
            \\
            
            {\color{wine}Event Module}: ~~~$\psi_{z\rightarrow z'},~p_{z\rightarrow z'}$

            \hfill
    \end{minipage}
    \hfill
    \begin{minipage}[h]{.50\linewidth}\small
        \centering
        \begin{tabular}{r|c}
            $z$ & latent mode (one--hot)\\\hline
            $X$ & collection of subtrajectories\\\hline
            $\{t_k\}$ & event times \\\hline
            $\theta$ & encoder parameters \\\hline
            $\omega$ & Neural ODE parameters\\\hline
            $F_z$ & mode--controlled Neural ODE \\\hline
            $\psi_{z\rightarrow z'}$ & jump networks \\\hline
            $p_{z\rightarrow z'}$ & normalizing flows
        \end{tabular}
    \end{minipage}
\end{CatchyBox}
\subsection{Gradient Pathologies}
We provide some theoretical insights on the phenomenon of gradient pathologies with the simple example of a one--dimensional linear hybrid system with two modes and one \textit{timed} jump,
\begin{equation}\label{eq:toy}
    \begin{aligned}
		\dot x_t &=\left\{
		\begin{matrix*}[l]
		 ax_t &t<\tau\\
		bx_t &t>=\tau\\
		\end{matrix*}\right.&~~t\neq\tau\\
		x_t^+ &= cx_t&~~t=\tau 
	\end{aligned}
\end{equation}
We let the system to evolve in a compact time domain $\bT=[0, 1]$ such that $\tau\in\bT$. Given an initial condition $x_0\in\R$, the solution $x_1$ can be obtained as follows
\begin{equation}
    \begin{aligned}
        x_\tau &= e^{a\tau}x_0 && 1.~\text{integrate 1st flow until $t=\tau$}\\
        x_\tau^+ &= cx_\tau = c e^{a\tau}x_0 && 2.~\text{at $t=\tau$ apply jump}\\
        x_1 &= e^{b(1 - \tau)}x_\tau^+ = ce^{b(1 - \tau)}e^{a\tau}x_0 = ce^{a\tau + b(1 - \tau)}x_0 && 3.~\text{integrate 2nd flow from $t=\tau$ to $t=1$}
    \end{aligned}
\end{equation}
Alternatively, we can compactly write the solution at any time $t\in\bT$ as
\begin{equation}\label{aeq:sol}
    x_t = 
	\left\{
	\begin{matrix*}
	e^{at}x_0 &t<\tau\\
	ce^{a\tau + b(t-\tau)}x_0&t\geq\tau\\
	\end{matrix*}\right.
\end{equation}
Using the previous equation we can compute the gradient of solutions w.r.t. the parameters $a,~b,~c,~\tau$. In particular, we have
\begin{equation}
    \begin{aligned}
    	\frac{\dd x_t}{\dd a} &= 
    	\left\{
    	\begin{matrix*}
    	t e^{at}x_0 &t<\tau\\
    	\tau ke^{a\tau + b(t-\tau)}x_0&t>\tau\\
    	\end{matrix*}\right.\\
    	\frac{\dd x_t}{\dd b} &=
    	\left\{
    	\begin{matrix*}
    	0 &t<\tau\\
    	(t-\tau) ke^{a\tau + b(t-\tau)}x_0&t>\tau\\
    	\end{matrix*}\right.\\
    	\frac{\dd x_t}{\dd c} &= 
    	\left\{
    	\begin{matrix*}
    	0 &t<\tau\\
    	e^{a\tau + b(t-\tau)}x_0&t>\tau\\
    	\end{matrix*}\right.\\
        \frac{\dd x_t}{\dd \tau} &= 
    	\left\{
    	\begin{matrix*}
    	0 &t<\tau\\
    	(a-b)ke^{a\tau + b(t-\tau)}x_0&t>\tau\\
    	\end{matrix*}\right.
        \end{aligned}
\end{equation}
Now let us consider a loss function computed on the mesh solution points of the trajectory
$$
    L = \sum_{k = 1}^K \gamma(x_{t_s}),\qquad0<t_1<\cdots<t_S<1,~t_s\neq\tau
$$
of which we wish to obtain the minimizers $a^*,~b^*,~c^*,~\tau^*$ via e.g. application of gradient descent methods. The gradient of the cost function w.r.t. any of the parameters $\theta \in \{ a, b, c, \tau\}$ is given by
$$
    \frac{\dd L}{\dd\theta} = \sum_{k=1}^S\frac{\dd\gamma(x_{t_s})}{\dd x}\frac{\dd x_{t_s}}{\dd \theta}.
$$
Simultaneous estimation of both the optimal dynamic parameters $a^*,~b^*,~c^*$ and a randomly initialized event time $\tau^*$, will result in gradients of certain parameters to vanish or be completely incorrect. 

Specifically, we note that parameter $\tau$ determines, beyond the specific time when the jump event occurs, also which parameters are responsible for computation of solution points $x_{t_s}$. Consider the following two scenarios, where mode $1$ is the first vector field of \eqref{eq:toy} and $2$ is the second (post--event):

\textbf{(1)} Initialization of $\tau$ is an over--estimation of $\tau^*$ at the beginning of training. If this is the case, for $t_s$ such that $\tau>t_s>\tau^*$ the mode is missclassified i.e. should be $2$, but is still $1$ due to the delayed event time $\tau$. The gradient w.r.t $b$ of loss computed on solution points $x_{t_k}, \tau>t_s>\tau^*$ is then wrongfully set to zero.

\textbf{(2)} $\tau$ is an under--estimation of $\tau^*$. The same reasoning applies, except that for $\tau^*>t_s>\tau$ the mode is misslassified to $2$, although it should be $1$. This, in turn, affects the gradients for $b$, which results different than $0$ despite the fact that $b$, from \eqref{eq:toy} should not be affecting the solution at points $t_s < \tau^*$. Any value taken by this gradient is thus incorrect, and can greatly affect the optimization procedure

We have shown how gradient pathologies exist even in the idealized linear case. In nonlinear systems with multiple events (including stochasticity) these effects can have a great empirical effect on a training procedure. The trajectory segmentation first approach of NHAs is designed to minimize their impact.
\subsection{Extensions and Limitations}
\paragraph{Automata reconstruction via symbolic regression}
NHAs with {\tt categorical} encoders recover either a representation using the minimum number of modes necessary, corresponding to those of the system, or can be pruned due to immunity to mode mixing (discussed in Appendix B.3).

This property allows application of \textit{symbolic regression} (SR) to reconstruct an analytic expression for each differential equation driving a system mode. This step grants domain experts a method to validate and certify the results, and enables construction of a human--readable automata representation for the SHS. 

Clustered SR results can be improved by leveraging the \textit{universal differential equation} approach employed in example by \cite{rackauckas2020universal} for unimodal differential equations, by utilizing the decoder $F_z$ as an \textit{interpolating} source of additional trajectory data for each mode.
\paragraph{Latent hybrid automata from observations}
Learning methods for dynamical systems often introduce structure in latent space to enable control and identification from raw observations \cite{mitchell2020first, zhong2020unsupervised}. Practical application for hybrid systems, such as robotic manipulation \cite{johnson2016hybrid}, locomotion \cite{holmes2006dynamics}, and traffic networks \cite{hespanha2004stochastic}, might benefit from learning models structurally equipped with latent NHAs.

Optimal design of NHA modules for latent applications remains a difficult open question, as the analysis of deep models with latent spaces designed to evolve in continuous--time is also in its 
infancy.
\paragraph{Unified benchmark for model development}
Despite the importance of hybrid systems in engineering, wide differences in techniques across domains have historically made it difficult to develop and preserve a unified set of benchmarks. 

Evidence from other deep learning disciplines e.g. computer vision highlights the importance of consistent and curated benchmark datasets to track and measure the impact architecture and method optimizations. We argue further benchmark design, along with larger datasets, to be a necessary step required to trigger an {\tt ImageNet}--like \cite{deng2009imagenet} moment for general neural differential equations and thus also NHAs. As an additional challenge, we note that performance of continuous neural models is in general highly impacted by the numerical method used for forward and backward inference, with optimal methods usually system or application dependent. This makes decoupling architecture improvements from the numerical underpinnings harder than for traditional models.

\subsection{Detailed Feature Comparisons with Related Work}
Table \ref{tab:feat_comp} compares the proposed method with recent learning based approaches in terms of features. We use \xmarkf~for features that are either absent or incompatible with a given method, or features that have not been tested or verified, although the method itself may be adapted to include it.

We consider the following:
\begin{itemize}[leftmargin=0.4in]
    \item \textbf{Recovery of flows and events:} can state--space vector fields be learned along with the events. In \cite{jia2019neural}, learned \textit{latent} dynamics aid in the intensity parametrization of the point process. State--space dynamics are not learned simultaneously with point process maximum likelihood training. \cite{chen2020learning} trains a neural network vector field along with a parametrized event function.
    \item \textbf{Stochastic events:} has the method been shown to be compatible with stochastic events. The formulation of \cite{chen2020learning} can parametrize stochastic events via inverse sampling, but no experiments have been performed, likely due to difficulties in learning stochastic events from a full trajectory.
    \item \textbf{Mode identification:} does the method recover modes of a multi--mode hybrid system, and can the vector field approximate a different dynamics for each. NHAs are the first method to tackle this setting.
    \item \textbf{Adaptive end--time:} can the method adjust event times by calculating gradients with respect to integration end--times. Core contribution of \cite{chen2020learning} is an implicit differentiation formulation to adapt end-times. While adaptive segmentation has been discussed as being compatible with NHAs, no targeted experiments on this technique have been carried out. The extension is left as future work.
    \item \textbf{Intensity--free parametrization:} does the method use intensity--free parametrizations to avoid numerically solving integrals to sample from next--event densities. \cite{jia2019neural,chen2020learning} parametrize the intensity as only a single mode is considered. NHA use normalizing flows to approximate these densities directly, since intensity parametrization scale poorly as the number of system modes increases.
\end{itemize}

\begin{table}[b]
    \centering
    \begin{tabular}{c|ccccc}
        Method & Recovery of & Stochastic & Mode & Adaptive & Intensity--free \\
        & flow $+$ events & events & identification & end--time & parametrization  \\
        \hline
        NJSDE \cite{jia2019neural} & \xmarkf & \checkmarkf & \xmarkf & \checkmarkf & \xmarkf  \\
        Neural Event ODE \cite{chen2020learning} & \checkmarkf & \xmarkf & \xmarkf & \checkmarkf & \xmarkf  \\
        {\color{blue!70!white} \textbf{NHA}} (this work) & \checkmarkf & \checkmarkf & \checkmarkf & \xmarkf & \checkmarkf 
    \end{tabular}
    \vspace{0.5mm}
    \caption{Feature comparison between neural models for hybrid systems. \xmarkf~is used to indicate features that are either not compatible, or have not been verified in the original work.}
    \label{tab:feat_comp}
\end{table}

\subsection{Broader Impact}
This work represents a first attempt in developing a data--driven, learning based technique for \textit{stochastic hybrid system} (SHS) identification and control. As discussed in the main text, existing methods currently rely on strict assumptions that severely limit their utilization in practice. Applications domain the SHS formalism provides an accurate language to describe a target system are most likely to be affected by the availability of NHAs as a method to improve partial mathematical models using data or construct from scratch a model of the system and its automata representation. The impact of NHAs here is thus likely to be similar in scope to the impact of neural differential equations in science and engineering.

We do not anticipate significant negative environmental impact from the adoption of NHAs as these models are still orders of magnitude smaller than other large deep learning architectures for domains such as natural language.

\section{Experimental Details}

\paragraph{Hardware and software resources} Experiments have been performed on a workstation equipped with a 48 threads \textsc{AMD Ryzen Threadripper 3960X}, a \textsc{NVIDIA GeForce RTX 3090} GPUs and two \textsc{NVIDIA RTX A6000}. All models and datasets fit in a single GPU. The software implementation of NHA leverages the $\tt PyTorch$ framework. ODE solvers and numerical methods for hybrid systems have been developed from scratch and are included in the submission. 
\paragraph{Common experimental settings} In all experiments, unless specified, the NHA mode encoder $\cE_\theta$ is capped with a ${\tt softmax}$ activation computing the probabilities of a categorical distribution from which then the one--hot mode $z$ is sampled. 

Gradients through the sampling operation are computed via \textit{straight--through--estimation} (STE) \cite{bengio2013estimating}, which can be implemented with a {\tt stop\_gradient} (e.g. {\tt detach} in {\tt PyTorch}) operation present in modern deep learning frameworks. %
Let $z$ be the one--hot representation of system mode, and $p$ a vector of probabilities for the corresponding {\tt categorical} distribution, computed as output of a neural network parametrized NHA encoder $\cE_\theta$. STE can be realized in a single line as $z - p.{\tt stop\_gradient}() + p$. STE ensures the output of the encoder $\cE_\theta$ is strictly one--hot encoded, while simultaneously ensuring that the gradients backpropagate directly through the probabilities. 

The data--controlled Neural ODE decoder $F_z$ incorporates mode information to select 
\[
\dot x = \sum_{i=1}^m z_i~f_i(t, x, \omega_i)
\]
where $z$ is one--hot encoded, and thus only a single neural network vector field $f_i$ with parameters $\omega_i$ determines the solution.

\subsection{Identification of Reno {\tt TCP}}
\paragraph{Experimental setup}
Tables \ref{table:noisy_clusterin22g} provide hyperparameters for data simulation and training of NHA. Unless otherwise specified simulation, training and testing is repeated for $10$ different random seeds.

The complete experiment on learning the Reno TCP system involves multiple stages: (i) mode recovery and (ii) training of the NHA event module. As baselines for (i), we consider several Neural ODE variants similar to NHA decoder $F_z$, with latent $z$ obtained in different ways. Latent Neural ODEs obtain $z$ as sampled for Normal distribution $\cN_\theta := \cN(\mu_\theta, \sigma_\theta)$ parametrized by a neural network matching the architecture of $\cE_\theta$. Reparametrization is used to backpropagate through the sampling procedure. \textit{Data--controlled Neural ODEs} (DC--NODEs) are comparable to Latent Neural ODEs model with the major difference in the computation of latents as $z = g_\theta(X)$ with $g_\theta$ once again matching $\cE_\theta$. All decoders $F_z$ are equivalent, except in the case of Augmented Neural ODEs (the Neural ODEs is zero--augmented \cite{massaroli2020dissecting} i.e. $z:=0$) where the absence of the encoder is balanced by a more expressive decoder with three--layers. The normalizing flows $\psi_{z\rightarrow z'}$ in NHA event modules are designed as spline flows \cite{durkan2019neural} with two layers. We use the standard implementation of spline flows in {\tt Pyro} \cite{bingham2019pyro}.

During mode recovery, all models are trained on 5--folds of $5$. Training is performed by parallel integration across subtrajectories $X_i$ using the {\tt Runge--Kutta}$4$ explicit solver. All gradients are computed via reverse--mode automatic differentiation. We test models with lowest cross--validation reconstruction MSE loss, since cross--validation v--score would not be available without ground--truth labels. We find that lowest reconstruction loss often correlated with best v--measure.
\begin{table}[t]
    \footnotesize
    \begin{tabular}{ll} \toprule
        \textbf{Simulation Hyperparameter} & \textbf{Value} \\ \midrule
        Number trajectories & $40$ \\
        ODE solver & {\tt Dormand--Prince} \\
        Tolerances (abs, rel, event) & $10^{-6}, 10^{-6}, 10^{-4}$ \\
        $\tau_{\tt off}$ & $3$ \\
        $\eta$ & $1$ \\
        $n_{\tt ack}$ & $2$ \\
        $p_{drop}$ & $0.05$ \\
        $\kappa$ & $4$ 
    \end{tabular} 
    \hfill
    \begin{tabular}{ll} \toprule
    \textbf{Training Hyperparameter} & \textbf{Value} \\ \midrule
    Training iterations (mode recovery) & $4000$ \\
    Encoder $\cE_\theta$ learning rate & $5\cdot 10^{-4}$ \\
    Decoder $F_z$ learning rate & $10^{-2}$ \\
    Optimizer & {\tt Adam} \\
    $\cE_\theta$ layer dimensions & $[\cdot, 64, 65, 64, m]$ \\
    $\cE_\theta$ activation & {\tt ReLU} \\
    $\cE_\theta$ dropout & $[0.3, 0.3, 0.3]$ \\
    $F_z$ layer dimensions & $[2 + m, 2]$ \\\midrule
    Training iterations (event training) & $4000$ \\
    Learning rate & $2\cdot 10^{-3}$ \\
    Optimizer & {\tt Adam} \\
    $\psi_{z\rightarrow z'}$ layer dimensions & $[2, 32, 2]$ 
\end{tabular} 
    \vspace{4mm}
	\caption{\footnotesize Hyperparameters of \textbf{[Left]} TCP data simulation \textbf{[Right]} NHA in the TCP experiment, mode recovery and event module training.}
	\label{table:noisy_clusterin22g}
\end{table}
Baselines for mode clustering (ii) include standard clustering algorithms k--means++ \cite{arthur2006k}, hierarchical \cite{murtagh2012algorithms} and DBSCAN \cite{birant2007st}. We use {\tt scikit--learn} \cite{pedregosa2011scikit} implementation of all baseline algorithms. We perform light hyperparameter tuning on ground--truth labels for k--means++ and hierarchical to optimize their performance in the range $m\in[3,5,10]$. DBSCAN is similarly tuned to optimize its performance with parameter $\epsilon$ indicating the maximum size of neighourhoods around a data point. Subtrajectories classified as noise by DBSCAN are counted as incorrectly clustered.

We observe DBSCAN performance to be correlated to NHA self--supervised mode recovery. Both methods excel when density of data points under some metric is indicative of cluster separation. However, NHA self--supervision relies on the additional inductive bias of data points in a subtrajectory representing observations of a solution of an ODE, whereas DBSCAN does not. The denser the trajectories, the more restricting the Neural ODE representation limitations, and the easier each cluster is to find. Due to the similarity in their working principle, DBSCAN performance can be used as a quick sanity check to determine whether the dataset is suitable to NHA mode recovery.
\subsection{Robustness to Segmentation Noise}
\begin{wraptable}{r}{0.6\textwidth}
    \footnotesize
    \begin{tabular}{llllll} \toprule
         &   & $\Delta$& \\ \midrule
        \textbf{Model} & $p = 0.1$ & $p = 0.3$ & $p = 0.5$ \\ \midrule
        k--means{\tt $++$} & {\color{red!20!black}$-0.01$} ($0.29$) & {\color{red!40!black}$-0.05$} ($0.25$) & {\color{red!55!black}$-0.08$} ($0.22$) \\
        hierarchical & {\color{red!4!black}$-0.00$} ($0.31$) & {\color{red!20!black}$-0.02$} ($0.29$) & {\color{red!30!black}$-0.03$} ($0.28$) \\
        DBSCAN & {\color{red!50!black}$-0.08$} ($0.61$) & {\color{red!70!black}$-0.28$} ($0.41$) & {\color{red!100!black}$-0.46$} ($0.23$) \\ \midrule
        NHA--3 & {\color{green!20!black}$+0.02$} ($0.88$) & {\color{red!60!black}$-0.13$} ($0.73$) & {\color{red!100!black}$-0.45$} ($0.41$) \\ 
        NHA--5 & {\color{red!1!black}$-0.00$} ($0.91$) & {\color{red!70!black}$-0.17$} ($0.74$) & {\color{red!100!black}$-0.48$} ($0.43$) \\ 
        NHA--10 & {\color{red!40!black}$-0.07$} ($0.89$) & {\color{red!60!black}$-0.25$} ($0.71$) & {\color{red!100!black}$-0.41$} ($0.45$) \\ 
    \end{tabular} 
	\caption{\footnotesize v--measure performance differences after self--supervised mode recovery performed on a dataset with imperfect segmentation noisy segmentation of the TCP dataset of the main text. We evaluate each method under an increasing degree of data corruption. Hyperparameter $p$ indicates the probability for a subtrajectory $X_i$ to be subject to a noisy segmentation i.e. to have the index determining its initial condition be perturbed and shifted either left (before) or right (after). In parenthesis, the v--measure of each model in a given noisy regime.}
	\label{table:noisy_clustering}
\end{wraptable}
We investigate robustness of NHA mode recovery and related baselines to a noisy segmentation in subtrajectories $X_i$. To simulate incorrect segmentations, we collect segmentation indices and perturb them by adding or removing an uniformly sampled from $[1, 10]$. Each index has a $p$ probability of being corrupted by noise, and we repeat mode recovery with $p\in[0.1,0.3,0.5]$ ($3$ times per $p$). Shifting left or right by values sampled from $[1, 10]$ results in significant data corruption; certain subtrajectories, being shorter than $10$ points, can be completely absorbed into a different subtrajectory. Table \ref{table:noisy_clustering} reports the differences in v--measure compared to the results of Section 5.1. For each baseline, we report best results in terms of hyperparameter $m\in[3,5,10]$, typically $m=10$, corresponding to $\epsilon=1$ for DBSCAN. Although specific perturbation patterns may affect the model more than others, the trend uncovered by the results is clear. 

\subsection{Switching Linear System and Mode Mixing}

\begin{figure}
    \centering
    \includegraphics{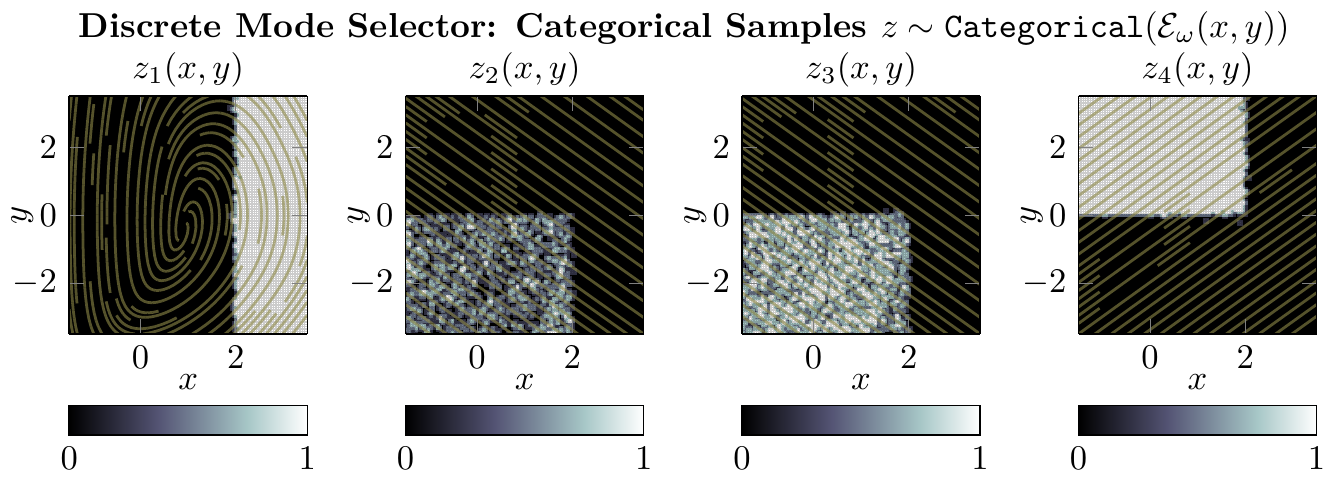}
    \caption{Reconstructed conditional vector fields $F_z$ and corresponding mode classification boundaries. The {\tt categorical} encoder uses two identical modes for a single ground--truth vector fields.}
    \label{fig:case2}
\end{figure}

\paragraph{Experimental setup}
We considered the two--dimensional switching linear system reported in \cite{chen2020neural}, described by the dynamics
\begin{equation}\label{eq:sls}
    (\dot x_t, \dot y_t) = f(x_t, y_t) := 
        \left\{
            \begin{aligned}
                (-y_t, x_t + 2) \quad& \text{ if }~ x_t\geq 2\\
                (-1, -1) \quad& \text{ if }~ x_t< 2\land y_t\geq 0\\
                (1, -1) \quad& \text{ if }~ x_t< 2\land y_t< 0
            \end{aligned}
        \right.
\end{equation}
We performed an ablation study on the effect of the categorical sampling for the mode selection in NHAs in presence of redundant "free" modes. In particular, we considered the following learning model
\begin{align}
    (\dot x_t, \dot y_t) = \sum_{i=1}^4 w^i_t f_{i}(x_t, y_t)
\end{align}
with one redundant mode. $F_i$ ($i=1,2,3,4$) was two-layers neural networks with 32 neurons each, {\tt softplus} activation on the first hidden layer and hyperbolic tangent activation on the second one. We then defined two variants of the model: a first variant with $w_t = (w^1_t, w^2_t, w^3_t, w^4_t)$ directly obtained via {\tt softmax} normalization of the output of a neural network $g$,
\[
    w_t = {\tt softmax}~g(x_t, y_t);
\]
and a second one where $w_t$ is obtained by a categorical sample conditioned by $g(x_t, y_t)$, i.e.
\[
    \forall t\quad w_t\sim {\tt categorical}({\tt softmax}~g(x_t, y_t))
\]
$g$ was fixed as a neural network made up by two layers with 64 units and {\tt SiLU} ({\tt swish}) activation.
The two models were trained on a $L_1$ reconstruction loss of nominal trajectories of the system \eqref{eq:sls}. We introduced a regularization term penalizing the squared error on un--normalized finite differences of nominal/reconstructed trajectories as a proxy for the vector field information.
\begin{wrapfigure}[17]{r}{0.365\textwidth}
    \centering
    \includegraphics[scale=0.6]{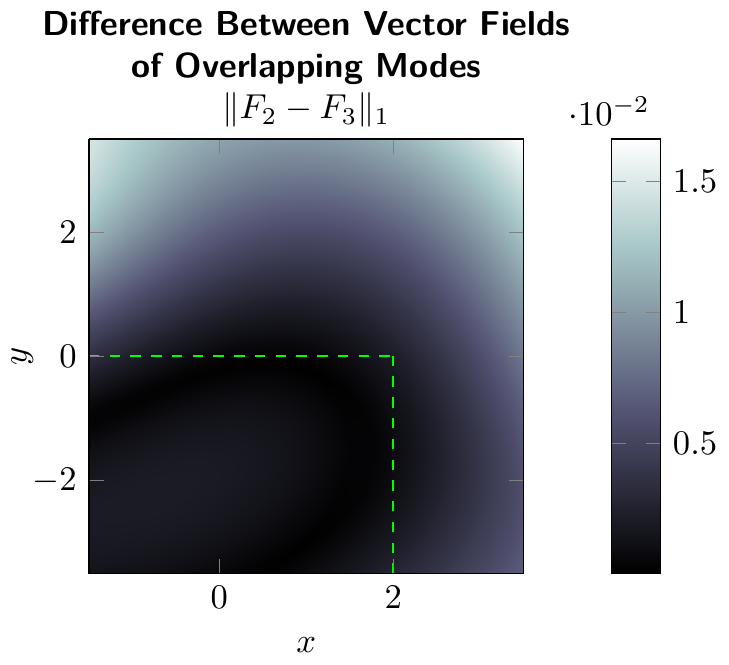}
    \caption{Similarity between learned mode vector fields $F_2$ and $F_3$ of Figure \ref{fig:case2}. The two vector fields are equivalent in the region of interest, as indicated by the L1--norm, and the corresponding modes can thus be merged.}
    \label{fig:case3}
\end{wrapfigure}
\paragraph{Mode pruning}
Uniqueness theorems for ODE solutions guarantee that, given an initial condition and a mode latent code $z$, the decoder $F_z$ will always produce the same trajectory. Immunity to mixing for categorical latents enables mode pruning and recovery of a minimal representation. If $\cL_r$ saturates, the encoder has not been initialized with a sufficient number of modes $m$. Redundant modes may be pruned, in example, by merging them if a similarity measure between the corresponding vector fields $F_{z_i},~F_{z_j}$ e.g. difference in a given norm calculated on data trajectories, is \textit{small enough}. Figure \ref{fig:case2} provides an example result of the second scenario discussed in Section 5.2, where {\tt categorical} NHA encoders use more than a single latent mode for a target underlying mode. However, due to their immunity to mode mixing, the vector fields are equivalent, and can be merged. We show this in Figure \ref{fig:case3}, where the L1--norm between $F_2$ and $F_3$ is shown to be small in the region where the corresponding modes $2$ and $3$ are active.

\subsection{End--To--End Learning of Hierarchical Controllers for Dynamical Systems}

\paragraph{Experimental setup}
Our objective is to control a swarm of differential drive robots moving on a planar space. The system dynamics are
\begin{equation}
    \begin{aligned}
    \dot x_1 &= u_v\cos{\theta}\\
    \dot x_2 &= u_v\sin{\theta}\\
    \dot \vartheta &= u_r
    \end{aligned}
\end{equation}
where $\vartheta$ is the orientation of a single robot. The control $[u_v, u_r] \in \R^2$ is obtained via the low--level, time--invariant feedback controller $u_z:=u(x_1, x_2, \vartheta, z)$, with $z$ produced and switched every $5$ seconds by the planner. Both the controller and the policy $\pi$, are parametrized by a neural network. We use {\tt Adam} for both networks, with learning rates $10^{-3}, 10^{-3}$. 

The training of low--level controller $u_z$ and high--level planner $\pi$ is carried out concurrently. We perform batch training on robot swarms of $N=80000$. At the beginning of each episode, we sample the initial conditions uniformly in a square region, with each robot rotated according to a random orientation also uniformly sampled in $[0, 2\pi]$. During each episode, we simulate the switching and control behaviour of each robot with respect to two pre-defined map layouts, both shown in the alternating resource pattern of the upper-left plot of Figure \ref{fig:ex3-a}. Each map layout has $M=5$ resource locations, where two auxiliary scalar variables $r_1$ and $r_2$ specify, for each resource, its planar location.

We train policy networks $\pi(x_1, x_2, r_{11}, r_{21}, ...,  r_{1m}, r_{2m})$ where the inputs correspond to the concatenated robot states, together with the flattened resource locations. The $M-$dimensional softmax output determines a categorical probability distribution over the resources is then used to sample a resource target.

A reward can be assigned to each robot based on the ability to select the correct target. For each robot, we generate a reward of 1 if the selected target is correct, and 0 otherwise. Let $G(s_i, p_i)$ be the reward of robot $i$ in the swarm, we can compute the reward for a swarm in a map by $R=\sum^{N}_{i}\frac{G(s_i, p_i)}{1000}$ where $s$ is the categorical sample from the distribution over the targets, and $p$ are the reference targets computed by $\arg\min_j || [r_{1j}, r_{2j}] - [x_1, x_2] || $ (note $R \in [0, 80]$ for any one episode), with $x_1, x_2$ being the robot location at the time of switching. The target selector is trained by minimizing a  $\cL_{\pi} = -\frac{1}{T}\sum_t^T ln(\pi(s^{(t)}_p)*R$), where T is the number of alternating maps, and $s^{(t)}_p$ denotes the concatenated robot location and resource map with respect to the t$^{th}$ map layout used for training.

The target selected by the planned informs low--level controller $u_z$ via the corresponding resource location. In particular, we provide as input to $u_z$ the coordinates $z:=[r_1, r_2]$ of the target chosen by $\pi$. This augmented state is used by $u_z$ to resolve the robot's dynamics and drive the swarm closest to their selected targets. We train $u_z$ by solving a continuous--time optimal control problem with a terminal RMSE loss between the state reached by the robot and the objective set by the policy planner. Here, we integrate the system using the adaptive--step {\tt DormandPrince} \cite{dormand1980family} solver with tolerances $10^{-3}, 10^{-3}$. 

\paragraph{Discussion of results}
Figure \ref{fig:ex3} shows the average reward and control loss of the robot swarm during training, with both trends converging after 4000 episodes. Figures \ref{fig:ex3-a} and \ref{fig:ex3-b} show the generated control of a randomly generated swarm of 100 robots on two new maps. In the first map, the targets consist on alternating patterns of the learned map layouts, generating a straight line pattern which correctly captures the greedy robot policy imposed. The second map consists instead of a new, unseen, map layout within the alternation. The trained model is capable of generalizing the planning and control strategy to account for the new map layout, by redirecting the robots onto their closest resource in a wave-like pattern. On the first tested map, the model achieves $99.8\%\pm0.8\%$ average target accuracy for the 100 robot tested batch.  On the second tested map, the model achieves $98.5\% \pm1.95\%$ average target accuracy for the 100 robot tested batch.
\section{Realization of NHAs}
\subsection{Software Implementation of Hybrid Integration}
We provide documented {\tt Python} pseudo--code for the hybrid system adaptive integration algorithm used for dataset generation. This function can handle hybrid systems with multiple modes and transitions. Each possible event requires its own callback function with {\tt check\_event} and {\tt jump\_map} methods. We provide an example of one such callback under {\tt odeint\_hybrid}.

\begin{mintedbox}{python}
def odeint_hybrid(vf, x, t_span, solver, callbacks, atol, rtol, event_tol):
    """ODE solver for hybrid systems with multiple events."""
    # initialize event state tracker, one boolean for each possible event 
    # (or edge in the automata representation of the SHS).
    event_states = [False for _ in range(len(callbacks))]
    dt = initial_step_size(f, k1, x, t, solver.order, atol, rtol)
    
    while t < t_span[-1]:
        # tentative step
        x_step, x_err = solver.step(vf, x, t, dt)
        
        # check whether any event 
        # has been triggered in the interval [t, t + dt]
        new_event_states = [cb.check_event(t + dt, x_step) 
                                       for cb in callbacks]
                                       
        # has any event state moved from `False' to `True' in [t, t + dt]?
    	triggered_events = sum([(zp != z) & (z == False)
    		for zp, z in zip(new_event_states, event_states)])
    	# if an event / mode transition has been triggered,
    	# find exact event time and state
    	if triggered_events > 0:
                x, t = root_find_event(max_iters, event_tol)
                
                # if there is a conflict and multiple events are triggered, 
                # takes always the one with smaller ID
                zp = min([i for i, ev in enumerate(new_event_states) 
                           if ev == True])
                           
                t = t + dt
                # save state and time BEFORE and AFTER jump
                sol.append(x)
                eval_times.append(t)
                
                # apply jump func.
                x = callbacks[zp].jump_map(t, x)
                
                sol.append(x)
                eval_times.append(t)
    	
    	# when there are no events,
    	# proceed as usual with adaptive integration
    	else:
    	    error_ratio = compute_error(x_step, x_err, atol, rtol)
    	    accept_step = error_ratio <= 1
    	    
    	    if accept_step:
                    t = t + dt
                    sol.append(x)
                    eval_times.append(t)
    			
    	    else:
    	        dt = adapt_step(dt, error_ratio, safety, 
    	                        min_factor, max_factor, order)
    	        
    return eval_times, sol
\end{mintedbox}

The callbacks are in the form:
    
\begin{mintedbox}{python}   
class EventCallback(nn.Module):
        super().__init__()

    def check_event(self, t, x):
        raise NotImplementedError

    def jump_map(self, t, x):
        raise NotImplementedError


class StochasticEventCallback(nn.Module):

        super().__init__()
        self.exponential = Exponential(1)

    def initialize(self, x0):
        # sample one `s' for each batch in x0, to identify events 
        # as described in Section 2. Exponential, instead of Uniform
        # is used to avoid `log' computations.
        # Should be sampled again after every event is triggered. 
        self.s = self.exponential.sample(x0.shape[:1])

    def check_event(self, t, x):
        raise NotImplementedError

    def jump_map(self, t, x):
        raise NotImplementedError
\end{mintedbox}

\end{document}